\documentclass[10pt]{article}
\usepackage{ijcai17}
\usepackage{times}
\usepackage{latexsym}
\usepackage{subfigure}
\usepackage{graphicx}
\usepackage{multirow}
\usepackage{url}
\usepackage{color,soul}
\definecolor{darkcyan}{rgb}{0.0, 0.55, 0.55}
\definecolor{bluem}{rgb}{0.29, 0.33, 0.13}
\usepackage[small]{caption}

\usepackage{times}

\title{Context-aware Path Ranking for Knowledge Base Completion}
\author{Sahisnu Mazumder,~~~Bing Liu\\ 
Department of Computer Science, 
University of Illinois at Chicago, USA\\
sahisnumazumder@gmail.com,~~~liub@cs.uic.edu}

\begin{document}

\maketitle

\begin{abstract}
Knowledge base (KB) completion aims to infer missing facts from existing ones in a KB. Among various approaches, path ranking (PR) algorithms have received increasing attention in recent years. PR algorithms enumerate paths between entity-pairs in a KB and use those paths as features to train a model for missing fact prediction. Due to their good performances and high model interpretability, several methods have been proposed. However, most existing methods suffer from scalability (high RAM consumption) and feature explosion (trains on an exponentially large number of features) problems. This paper proposes a \textit{Context-aware Path Ranking} (C-PR) algorithm to solve these problems by introducing a selective path exploration strategy. C-PR learns global semantics of entities in the KB using word embedding and leverages the knowledge of entity semantics to enumerate contextually relevant paths using bidirectional random walk. Experimental results on three large KBs show that the path features (fewer in number) discovered by C-PR not only improve 
predictive performance but also are more interpretable than existing baselines.
\end{abstract}

\section{Introduction}
Knowledge bases (KBs), like Freebase \cite{bollacker2008freebase}, WordNet \cite{miller1995wordnet} and ConceptNet \cite{speer2012representing} have recently grown their popularity due to their applications in improving many NLP tasks like relation extraction \cite{44316}, syntactic parsing \cite{gesmundo2014projecting} and question answering \cite{berant2013semantic}. 
Such KBs contain large collections of facts about things, people and places often represented as triples, e.g., (\textit{Obama}, \textit{PresidentOf}, \textit{USA}). However, KBs are typically missing a large percentage of facts about common entities \cite{west2014knowledge} which makes them difficult to fulfill their full potentials. But, enriching KBs with all possible facts manually is also infeasible. Thus, researchers have devised techniques to automatically infer new facts from existing ones, which is formally known as the KB completion problem.

Broadly speaking, the main approaches to solving the KB completion problem include path ranking (PR) \cite{lao2011random,lao2015learning,gardner2014incorporating}; embedding  \cite{bordes2011learning,bordes2013translating,nickel2011three}; graphical models like Markov logic networks
\cite{jiang2012learning} and tensor/matrix factorization \cite{nickel2015review} methods. This paper focuses on path ranking (PR). In PR-based methods, KB is encoded as a multi-relation graph formed with a set of triples, where two nodes (entity-pair in a triple) are linked by a labeled edge (relationship in that triple). A path between two entities is a sequence of relations linking them. Given a relation $r$, PR-based methods enumerate paths (except path $r$) between entity-pairs linked by $r$ and use them as features to train a binary classifier to predict if new instances (entity-pairs) can be linked by $r$ or not.

Due to the high interpretability (as opposed to embedding techniques), no need for external logic rules (as opposed to MLN) \cite{wang2016knowledge} and better performance than latent features  \cite{toutanova2015observed}, PR-based algorithms have received increased attention in recent years. A number of works \cite{gardner2013improving,gardner2014incorporating,gardner2015efficient,lao2015learning} have proposed improved techniques. 
However, they still suffer from two major problems \textit{considering limitations in computational resources}: 
\begin{itemize}
	\vspace{-0.05cm}
	\item \textbf{Scalability}. Exploring all possible paths between entity-pairs using breadth-first search (BFS) with Subgraph Feature Extraction (SFE) \cite{gardner2015efficient} has shown state-of-the-art predictive performance compared to the random walk based PR approach \cite{lao2011random}. However, it is not always possible to have enough RAM to run the algorithm successfully on large graphs. Specifically, as the graph becomes \textit{denser} (degree of nodes increases), even BFS with SFE becomes intractable in moderate sized graphs (see Table 3). In such cases, we have to perform random walks. However, training models on millions of path features obtained via billions of random walks \cite{kyrola2013drunkardmob} is also not feasible with \textit{limited RAM}. And with disk based data structures, the training could get extremely slow.
	\vspace{-0.05cm}
	\item \textbf{Feature Explosion}. Even if we have sufficient RAM, 
	with a large number of paths (features), the length of the feature vector grows exponentially. Specially, if we want to derive new features from existing ones and augment them with the old ones like path-bigrams \cite{neelakantan2015compositional}, ``Any-Relation'' features \cite{gardner2015efficient} etc., the feature explosion problem makes the optimization (hence, training) computationally intensive and time consuming (see Table 3).
	\vspace{-0.05cm}
\end{itemize}
In such scenarios, we need a solution that can learn from \textit{fewer features} without suffering in predictive performance.

\begin{figure}
	\centering
	\includegraphics[height=5.1cm]{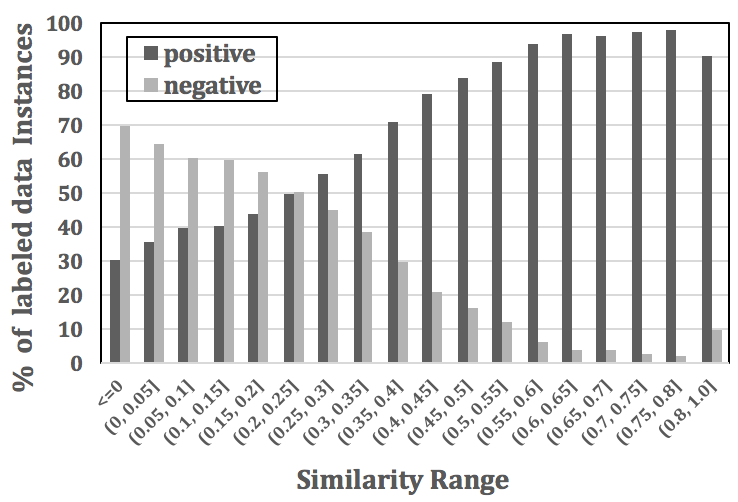}
	\vspace{-0.25cm}
	\caption{Correlation between contextual similarity and \% of data instance labels.}
	\label{task_example}
	\vspace{-0.25cm}
\end{figure}

A natural solution is to use context-aware path search, i.e., to find paths that are \textit{contextually relevant} with respect to the source and target entity pair. A path is considered as \textit{contextually relevant} if \textit{all}  intermediate entities encountered while following the path from source to target, are contextually relevant with regard to both the source and target entities. For example, given a source-target entity-pair (\textit{apple}, \textit{eat}), entities like ``\textit{fruit}'' and ``\textit{bite}'' are  contextually relevant to (\textit{apple}, \textit{eat}), while entities like ``\textit{company}'' and ``\textit{drink}'' are not, although they may be neighboring nodes of either ``\textit{apple}'' (`\textit{apple}', `\textit{IsA}', `\textit{company}') or ``\textit{eat}'' (`\textit{eat}', `\textit{RelatedTo}', `\textit{drink}') in the graph. Thus, given (\textit{apple}, \textit{eat}), choosing paths involving nodes like ``\textit{fruit}'', ``\textit{bite}'' etc., are more reasonable than those involving ``\textit{company}'' and ``\textit{drink}.'' Adopting this context-aware strategy in PR helps select more \textit{relevant} and \textit{discriminative} path features. To investigate the potential of this idea, we experimented with a randomly chosen set of labeled data instances (entity-pairs) from our Freebase dataset (see Sec 5.2). Figure 1 shows the correlation between contextual similarity of an entity-pair and existence (denoted by labels) of a relation in between them. We found that contextual similarity (computed using word vectors) is significantly correlated [$R^2=0.91$ (+ve label) and $R^2=0.85$ (-ve label) with exponential fit, where $R^2$ denotes the coefficient of determination] with the labels of data instances. Here (Figure 1), we see that contextual similarity can discriminate the positively labeled (relation exists) entity-pairs from negative ones (relation doesn't exist). The findings suggest that leveraging context-aware path search in PR-based methods might result in \textit{significant gain} in predictive performance with significantly \textit{reduced} number of features. 

In this paper, we propose a novel PR-based KB completion approach, namely \textit{Context-aware Path Ranking} (C-PR in short), which incorporates the aforementioned idea. In particular, C-PR learns the semantics of entities in KB using a neural word embedding model \cite{mikolov2013efficient} and leverages the contextual similarity information of entities for finding contextually relevant paths while performing random walks through the graph. Paths found by C-PR are more discriminative in nature and fewer in number compared to that found using straightforward random walk in existing PR-based methods. Details of C-PR is discussed in Sec 4.

To the best of our knowledge, this is the first study that extends PR by incorporating the contextual information of entity-pairs during path search. We empirically verify the effectiveness of C-PR by comparing it with the state-of-the-art baselines on three real-world, large-scale KBs with varying size and density: \textit{Freebase}, \textit{WordNet} and \textit{ConceptNet}. 
Experimental results show that C-PR substantially outperforms the baselines in terms of scalability, predictive performance and model interpretability.

\section{Related Work}
\label{sec:RW}
As discussed earlier, there are four main approaches to KB completion: path ranking,  embedding, graphical models and tensor/matrix factorization. C-PR is a PR-based algorithm. Thus, we focus on reviewing the related works in PR. 

PR is a random walk-based inference technique for KB completion. It was first proposed by \cite{lao2010relational} and later, extended by many successive works, including improving KB inference using latent syntactic cues 
\cite{gardner2013improving}, 
incorporating vector space similarity into random walk \cite{gardner2014incorporating}, introducing subgraph feature extraction \cite{gardner2015efficient} to generate more predictive paths and learning relational features with bidirectional random walk \cite{lao2015learning}. Other recent works include building predictive models with multi-task learning \cite{wang2016knowledge} and learning compositional vector space models \cite{neelakantan2015compositional,das2016chains}. However, none of these techniques uses 
entity's contextual information to find more relevant and predictive paths as C-PR.

Recently, \cite{freitas2014distributional} modeled entity semantics using term co-occurrence patterns 
and proposed an algorithm (called DNA) to find paths 
between two terms using a similarity threshold and maximum path length. One key difference between DNA and C-PR is that C-PR is based on word embedding while DNA is based on the traditional vector space model. Numerous papers \cite{baroni2014don,levy2015improving} have shown that word embedding is more powerful in representing word semantics. C-PR also differs from DNA in a number of other ways:  

(1) DNA tells how two entities can be linked through a 
related node sequence, whereas C-PR predicts missing relations between entities. (2) DNA uses unidirectional navigation and appends the target as the last node in the path when it reaches the maximum path length, which often do not ensure that the $2^{nd}$ last node and target actually have a direct relationship in KB. C-PR follows edges in KB to find complete paths between two nodes.  (3) DNA selects nodes related to \textit{only} the target node. So, they often go out of context with respect to the source.  
C-PR chooses nodes that are semantically related to \textit{both} source and target and automatically restrains the path exploration to go out-of-context. (4) If we increase the maximum path length, the accuracy of DNA drops as it explores less coherent paths (see Sec 6.3 of \cite{freitas2014distributional}). C-PR automatically constrains itself from going out of context. So, it does not explore longer irrelevant paths. (5) C-PR doesn\mbox{'}t use any fixed similarity threshold like DNA. It dynamically sets the threshold for each entity-pair. 

\section{Path Ranking Algorithm}
\label{PRA}
The basic idea of a path ranking (PR) algorithm is to enumerate paths between two entities in a KB and use those paths as features to train a model for predicting missing relations between them. Here, a path is a sequence of relations $\langle  r_1, r_2, .., r_l \rangle$ that links two entities. A typical PR algorithm works as follows: Given an entity pair ($h$, $t$), PR performs random walks over KB (encoded as a multi-relation graph) to find paths starting from $h$ (or $t$) and ending at $t$ (or $h$) with bounded length. Next, PR selects a set of paths as features, according to precision-recall measure \cite{lao2011random} or simply frequency \cite{gardner2014incorporating}. Once the features are selected, given an entity pair ($h$, $t$) and a path $\pi$, PR computes random walk probability $P$($t \mid h$, $\pi$) \cite{lao2011random} (the probability of arriving at $t$ given a random walk starting from $h$ and following exactly all relations in $\pi$) and uses the probability as the feature value for path-feature $\pi$.  Finally, PR constructs a training data $D_r$ for each relation $r$ by combining a set of positively labeled (linked by $r$) and a set of negatively labeled (not linked by $r$) entity-pairs. Next, PR trains a binary classifier for $r$ on the feature matrix of $D_r$ constructed using the path features extracted for each entity pair ($h$, $t$) $\in D_r$. The trained classifier for relation $r$ is then used to predict whether a given test entity pair (which is not presently linked by $r$ in KB) can be linked by $r$ or not.

\section{Proposed Context-aware Path Ranking}
\label{C-PR}
Given a KB containing a collection of triples $\mathcal{T}$=\{($h$, $r$, $t$) $\mid h$, $t \in E$, $r \in R$ \}, where $E$ is the entity set and $R$ is the relation set, we encode the KB as a multi-relation graph $G=(E, T)$ with entities represented as nodes and triples ($h$, $r$, $t$) $\in \mathcal{T}$ as directed edges labeled with $r$ from $h$ to $t$. Then, the KB completion problem can be defined as a binary classification problem, where given a relation $r \in R$ and an entity pair ($h$, $t$) such that ($h$, $r$, $t$) $\notin \mathcal{T}$, we want to predict whether $h$ and $t$ can be linked by $r$ or not. In PR-based KB completion approaches, each relation $r$ is associated with a set of labeled training instances formed by extracting path features (Sec \ref{PRA}). C-PR proposes a novel random walk-based path feature extraction technique, by leveraging the context of the source-target entity-pair while exploiting the graph structure. The path feature extraction of C-PR works in two steps: \textit{contextual similarity computation} and \textit{context-aware path finding}.  

To compute \textit{contextual similarity}, we use the distributed semantics of entities in $G$ using neural word embedding \cite{mikolov2013efficient,mikolov2013distributed}. Given a large corpus $\mathcal{C}$, 
a word embedding model learns the embedding of words 
in a continuous vector space to
capture the distributed semantics of words based on their contexts in $\mathcal{C}$. We use the word2vec model in \cite{mikolov2013efficient} 
with the skip-gram architecture. 
C-PR currently employs a pre-trained word2vec model for Freebase\footnote{https://code.google.com/archive/p/word2vec} [with 1.4M vocabulary size and 1000D word vector for each Freebase entity] and a pre-trained Google News word2vec$^1$ model [containing 300D vectors for 3M words and phrases] (for WordNet and ConceptNet experiments\footnote{We found high coverage of WordNet and ConceptNet entities in the vocabulary of Google News word2vec model.})
both trained on Google News Corpus (size 100B). Given the word vectors $W_h$ and $W_t$ for an entity pair ($h$, $t$) in $G$, we compute the contextual similarity of $h$ and $t$ as $sim(h, t) = \frac{W_h \cdot W_t}{\| W_h \| \| W_t \|}$ and use the measure for navigating through the graph structure for finding contextual paths.

Given the source node $h$ and target node $t$ in $G$, \textit{Context-aware Path Finding} enumerates paths between $h$ and $t$ using a bidirectional random walk (RW in short). The bidirectional RW performs random walk starting from both $h$ and $t$ simultaneously and meets in some intermediate common node reached from both ends. Let $v_f^k$ be the last node visited from the forward direction and $Fseq$=$\langle  h, v_f^1, v_f^2, …, v_f^k \rangle$ be the forward node sequence enumerated so far in a given bidirectional RW considering source as $h$ and target as $t$. Similarly, let $v_b^k$ be the last node visited from the backward direction and $Bseq$=$\langle t, v_b^1, v_b^2, …, v_b^k \rangle$ be the backward node sequence enumerated so far in the same random walk. Then, for a given random step in forward direction, in order to expand the forward sequence $Fseq$, the \textit{Context-aware Path Finding} performs the following steps:

\begin{table*}[th]
\small
\centering
\caption{\small Example of C-PR path search execution trace on a given KB Completion Query from our ConceptNet dataset.}
\label{my-label}
\begin{tabular}{|c|l|}
\hline
KB Completion Query                                                                                                                                     & \multicolumn{1}{c|}{C-PR's Context-aware Path Search Execution Trace Example}                                                                                                                                                                \\ \hline
\multirow{2}{*}{\begin{tabular}[c]{@{}c@{}}Entity-pair: \textbf{(oven, burn)} \\ Query Relation: \textbf{UsedFor} \\Source-Target \\ Similarity: 0.2039\\ \end{tabular}} & \begin{tabular}[c]{@{}l@{}}After iter-1: Fseq: {[}`oven', `rack'{]}; Bseq: {[}`burn', `wood'{]},\\ During iter-2: Fseq: {[}`oven', `rack', `grill'{]}, Backward Seq: {[}`burn', `wood', \textcolor{blue}{`pyrography'}{]}. \\ Bseq went out-of-context at `pyrography' with $\Delta (pyrography, wood)$=-0.044\textless 0. path expansion stopped\end{tabular}                                                                                                                                                                                                            \\ \cline{2-2} 
                                                                                                                                                 & \begin{tabular}[c]{@{}l@{}}After iter-1: Fseq: {[}`oven', `microwave'{]}; Bseq: {[}`burn', `smoke'{]},\\ During iter-2: Fseq: {[}\textcolor{darkcyan}{`oven'}, `microwave', `cook'{]}; Bseq: {[}`burn', `smoke', \textcolor{darkcyan}{`oven'}{]},\\ Node overlapping detected during Bseq expansion at `oven',  Pseq: {[}`oven', `smoke', `burn'{]} \\ Full path (N: Node, R: Relation): `oven(N)'-- AtLocation(R)$^{-1}\rightarrow$  `smoke(N)'-- RelatedTo(R)$\rightarrow$ `burn(N)'\\ Inferred Path feature for the given query from current random walk: ``\textbf{\textcolor{bluem}{AtLocation$^{-1}\rightarrow$ RelatedTo}}''\end{tabular} \\ \hline
\end{tabular}
\vspace{-0.35cm}
\normalsize
\end{table*}

\textbf{Step-1.} The Path Finding algorithm finds the set of contextual neighbors 
of $v_f^k$. For a node $v$ in $G$, its set of contextual neighbors  $N_{context}(v)$ is given by
$N_{context}(v)$ = \{$v^{'} \mid v^{'} \in N_{G}(v)$,  $\mathcal{R}elv(v^{'},h,t) \geq sim(h,t)$,  $v^{'}  \notin Fseq$\}, where $N_{G}(v)$ is the set of all neighboring nodes of $v$. $\mathcal{R}elv(v,h,t)$ is the contextual relevance of $v$ with respect to $h$ and $t$, computed as the linear combination of $sim(v, h)$ and $sim(v, t)$: $\mathcal{R}elv(v,h,t)=  \theta \times sim(v, h) + (1 - \theta) \times sim(v, t)$,~ 0 $\leq \theta \leq$ 1. $\mathcal{R}elv(v,h,t)$ measures how strongly node (word) $v$ is contextually related to both $h$ and $t$ considering the corpus $\mathcal{C}$. Currently, C-PR uses $\theta = 0.5$ to give equal importance to both source and target. If $\mathcal{R}elv(v,h,t) \geq sim(h,t)$, $v$ is considered as \textit{contextually relevant} to node pair ($h$, $t$); otherwise $v$ is excluded from $N_{context}(v)$ and does not take part in expansion of any node sequence further. 

Once $N_{context}(v_f^k)$ is computed, the Path Finding algorithm randomly selects a node $v \in N_{context}(v_f^k)$ and performs step-2. If $N_{context}(v_f^k)$ is $\phi$, it sets a flag indicating forward node sequence expansion process is stuck. If both forward and backward node sequence expansion process gets stuck, the algorithm returns an empty path.

\textbf{Step-2.}  The Path Finding algorithm checks whether $v$ = $v_b^l \in Bseq$, $l \leq k$ or not. If not, it performs step-3. If yes, it infers that it has found a complete path and stops the random walk from exploring nodes further. Next, the algorithm merges the forward node sequence $Fseq$ and reverse of backward node subsequence $Bseq[:l-1] = \langle t, v_b^1, v_b^2, . . ., v_b^{l-1}  \rangle$ to generate complete path node sequence  $Pseq = \langle h, v_f^1, . . . , v, v_b^{l-1}, . . ., v_b^1, t \rangle$. Finally, the sequence of relations is inferred from $Pseq$ by considering all the relations in between consecutive nodes in $Pseq$ and returned as the explored path in the given bidirectional RW.

\textbf{Step-3.}  If $v \notin Bseq$, the Path Finding algorithm computes the difference in contextual relevance of node $v$ and the last node ($v_f^{k}$) in $Fseq$ as $\Delta (v, v_f^k) = \mathcal{R}elv(v,h,t) - \mathcal{R}elv(v_f^k,h,t)$.  If $\Delta (v, v_f^k) \geq 0 $\footnote{We analyzed a set of path search results for a collection of entity-pairs in our dataset and found that whenever $\Delta (v, v_f^k)~$ \textless$~ 0$, node $v$ becomes more generic (e.g., generic words like `people', `animal' which may appear in many contexts) and often leads the path to become contextually irrelevant. C-PR leverages this idea and automatically restrains itself from going out-of-context.}, $v$ is regarded as ``\textit{in-context}'' for the nodes in $Fseq$ and appended as $v_f^{k+1}$ in $Fseq$. Otherwise, $v$ is inferred as \textit{out-of-context} (node that is generic and often appears in multiple semantically different contexts) with regard to the nodes in $Fseq$ and hence, excluded from further consideration for expansion of $Fseq$.

Similar steps (1, 2 and 3) as mentioned above are followed for expansion of $Bseq$ using simultaneous random walk from the backward direction. The overall iterative process (forward and backward node sequence exploration) in Context-aware Path Finding stops when the length of the enumerated path at a given iteration reaches a pre-defined maximum allowed path-length denoted as $\eta$. We set $numWalkers$ (number of random walks to be performed per source-target pair) as $n$. Table 1 shows an example of execution trace of C-PR.  

Once we enumerate a set of paths between an entity pair ($h$, $t$), we consider the paths as features following the typical PR method (Sec \ref{PRA}). But, instead of using random walk probabilities, we use binarized path feature vectors following  \cite{gardner2015efficient} which showed that using binary feature values (indicating the presence or absence of a path) instead of computationally expensive random walk probabilities gives statistically indistinguishable performance. 

Next, we train a binary classifier using logistic regression (LR) on training feature matrix and use it to predict test instances (see Sec \ref{sec:exp_setup} for labeled dataset generation). We use sklearn python ML library for training the LR model with L2-regularization. Other parameters of the LR model 
are: tolerance set to 0.0001, maximum iterations for convergence to 200, and class weight set as `\textit{balanced}'. We choose LR because LR not only has been used in existing PR-based approaches \cite{lao2010relational,gardner2014incorporating,gardner2015efficient}, but also has been shown to give better performances compared to SVM \cite{wang2016knowledge}.

\begin{table}[t]
\small
	\centering
	\caption{\small Statistics of our dataset}
	\label{my-label}
	\begin{tabular}{|l|l|l|l|}
		\hline
		& Freebase & Wordnet & ConceptNet \\ \hline
		\#Relations                                                                                 & 1,345    & 18      &     55       \\ 
		\#Entities                                                                                  & 13,871 & 13,595  &   88,931            \\ 
		\#Triples                                                                                   & 819,962 & 113,040 &   1.79M         \\  \hline
		\begin{tabular}[c]{@{}l@{}}\#Relations Tested\end{tabular}                               & 25       & 18      &     19       \\ \hline
		\begin{tabular}[c]{@{}l@{}}Avg. \# training\\ instances per relation \end{tabular}          & 3,970    & 2,188   &   4,000         \\ \hline
		\begin{tabular}[c]{@{}l@{}}Avg. \# testing \\ instances per relation \end{tabular}          & 998      & 544   &     1,000       \\ \hline
	\end{tabular}
	\normalsize
	\vspace{-0.1cm}
\end{table}

\begin{table*}[th]
\tiny
\centering
\caption{\small Comparison of Scalability of C-PR with the baseline methods [ $\eta$ = 7, $numWalker$ = 20 ].}
\label{my-label}
\begin{tabular}{|l|c|c|c|c|c|c|cc|c|c|}
\hline
                                                                                        & \multicolumn{8}{c|}{Baselines}                                                                                                                                                                                                                                                                                                                                                           & \multicolumn{2}{c|}{Our Method}                                                   \\ \hline
                                                                                        & B-PR                                         & B-PR+Bi                                      & BB-PR                                  & BB-PR+Bi                              & SFE-PR                            & \multicolumn{1}{l|}{SFE-PR+AR}                          & \multicolumn{1}{l|}{DNA-PR}                       & \multicolumn{1}{l|}{DNA-PR+Bi}                  & C-PR                                    & C-PR+Bi                                           \\ \hline
\multicolumn{1}{|c|}{\begin{tabular}[c]{@{}c@{}}Path Finding \\ Algorithm\end{tabular}} & \multicolumn{2}{c|}{\begin{tabular}[c]{@{}c@{}}Unidirectional RW\\ (Two-sided )\end{tabular}} & \multicolumn{2}{c|}{\begin{tabular}[c]{@{}c@{}}Bidirectional\\ RW\end{tabular}} & \multicolumn{2}{c|}{\begin{tabular}[c]{@{}c@{}}Breadth First Search\\ (Exhustive)\end{tabular}} & \multicolumn{2}{c|}{\begin{tabular}[c]{@{}c@{}}DNA + Unidirectional \\ RW (Two-sided )\end{tabular}} & \multicolumn{2}{c|}{\begin{tabular}[c]{@{}c@{}}Context-aware \\ Bidirectional RW\end{tabular}} \\ \hline
\multicolumn{11}{|c|}{Freebase ( in-memory word2vec model = 5.4GB,~~ in-memory Graph = 49 MB)}                                                                                                                                                                                                                                                                                                                                                                                                                                                                                              \\ \hline
\#Avg. Feat./rel.                                                                       & 458                                           & 1193                                          & 1815                                    & 4020                                   & 551M                                & 893M*                                                      & \multicolumn{1}{c|}{868}                          & 2084                                            & 834                                       & 1536                                                \\ \hline
RAM Consump.                                                                            & \multicolumn{4}{c|}{1.1 GB}                                                                                                                                                      & \multicolumn{2}{c|}{\textgreater\textgreater 100 GB}                                            & \multicolumn{2}{c}{7.2 GB}                                                                          & \multicolumn{2}{c|}{7.2 GB}                                                                     \\ \hline
Avg. tr. time/rel.                                                                      & 0.1576 sec                                    & 0.2559 sec                                    & 0.2994 sec                              & 0.3921 sec                             & 70.23 min                           & 105.64 min*                                                & \multicolumn{1}{c|}{0.167 sec}                    & 0.304 sec                                       & 0.1773 sec                                & 0.2339 sec                                          \\ \hline
\multicolumn{11}{|c|}{WordNet ( in-memory word2vec model = 3.4GB,~~ in-memory Graph = 7.1MB)}                                                                                                                                                                                                                                                                                                                                                                                                                                                                                               \\ \hline
\#Avg. Feat./rel.                                                                       & 164                                           & 258                                           & 358                                     & 506                                    & 2.5M                                & 6.2M                                                      & \multicolumn{1}{c|}{212}                          & 319                                             & 149                                       & 216                                                 \\ \hline
RAM Consump.                                                                            & \multicolumn{4}{c|}{376 MB}                                                                                                                                                      & 48GB                                & 49GB                                                      & \multicolumn{2}{c|}{5.3 GB}                                                                         & \multicolumn{2}{c|}{5.3 GB}                                                                     \\ \hline
Avg. tr. time/rel.                                                                      & 0.0724 sec                                    & 0.0900 sec                                    & 0.1138 sec                              & 0.1305 sec                             & 4.5 sec                             & 12.22 sec                                                 & \multicolumn{1}{c|}{0.0739 sec}                   & 0.0993 sec                                      & 0.0716 sec                                & 0.0882 sec                                          \\ \hline
\multicolumn{11}{|c|}{ConceptNet ( in-memory word2vec model = 3.4GB,~~ in-memory Graph = 136MB)}                                                                                                                                                                                                                                                                                                                                                                                                                                                                                          \\ \hline
\#Avg. Feat./rel.                                                                       & 221                                           & 386                                           & 556                                     & 858                                    & 386M                                & 716M*                                                      & \multicolumn{1}{c|}{339}                          & 555                                             & 517                                       & 719                                                 \\ \hline
RAM Consump.                                                                            & \multicolumn{4}{c|}{2.2 GB}                                                                                                                                                      & \multicolumn{2}{c|}{\textgreater\textgreater 100GB}                                             & \multicolumn{2}{c|}{6.2 GB}                                                                         & \multicolumn{2}{c|}{6.2 GB}                                                                     \\ \hline
Avg. tr. time/rel.                                                                      & 0.1401 sec                                    & 0.1750 sec                                    & 0.1723 sec                              & 0.2062 sec                             & 70.15 min                           & 109.82 min*                                                & \multicolumn{1}{c|}{0.1580 sec}                   & 0.1833 sec                                      & 0.1729 sec                                & 0.1948 sec                                          \\ \hline
\end{tabular}
\end{table*}

\begin{table*}[th]
	\small
	\centering
	\caption{\small Comparison of predictive performance of C-PR with the baseline methods [ $\eta$ = 7, $numWalker$ = 20 ].}
	\label{my-label}
	\begin{tabular}{|l|c|c|c|c|c|c|c|c|c|}
		\hline
		& \multicolumn{3}{c|}{Freebase}      & \multicolumn{3}{c|}{WordNet}       & \multicolumn{3}{c|}{ConceptNet}    \\ \hline
		\multicolumn{1}{|c|}{Model} & MAP    & Avg. F1 (+) & Avg. F1 (-) & MAP    & Avg. F1 (+) & Avg. F1 (-) & MAP    & Avg. F1 (+) & Avg. F1 (-) \\ \hline
		B-PR                       & 0.6747   & 0.4629     & 0.9069      & 0.8823 &  0.8083     & 0.9615     & 0.6298  &  0.3021    & 0.9068     \\ \hline
		B-PR+Bi                    & 0.6800   & 0.4710     & 0.9014      & 0.8834 &  0.8093     & 0.9606     & 0.6421  &  0.3431    & 0.9057  \\ \hline
		BB-PR                      & 0.7695   & 0.6309     & 0.9175      & 0.9409 &  0.9046     & 0.9781     & 0.6771  &  0.4235    & 0.9109      \\ \hline
		BB-PR+Bi                   & 0.7635   & 0.6057     & 0.9052      & 0.9283 &  0.8813     & 0.9709     & 0.6792  &  0.4537    & 0.9005     \\ \hline
		SFE-PR                     & 0.5579   & 0.1988     & 0.8261      & 0.8437 &  0.7298     & 0.9495     & 0.5888  &  0.2574    & 0.8214     \\ \hline
		SFE-PR+AR                  & 0.6079*  & 0.3975*    & 0.8498*     & 0.8820 &  0.8293     & 0.9599     & 0.7222* &  0.3657*   & 0.8641*     \\ \hline 
		DNA-PR                     & 0.6947   & 0.5079     & 0.9085      & 0.8916 &  0.8252     & 0.9646     & 0.6422  &  0.3333    & 0.9079      \\ \hline
		DNA-PR+Bi                  & 0.7057   & 0.5207     & 0.9032      & 0.8927 &  0.8208     & 0.9622     & 0.6574  &  0.3742    & 0.9059      \\ \hline \hline
		
		\multirow{ 2}{*}{C-PR}  & \textbf{0.7999} & \textbf{0.7105}      & \textbf{0.9292}     
		& \multirow{ 2}{*}{\textbf{0.9445}} & \multirow{ 2}{*}{\textbf{0.9082}}   & \multirow{ 2}{*}{\textbf{0.9792}}   & \textbf{0.7401}  & 0.5551 & \textbf{0.9221}     \\
		& \textbf{0.7996*} &  \textbf{0.7133*} & \textbf{0.9272*} &   &   &     & \textbf{0.7529}*       & 0.5776*    & \textbf{0.9253*}   \\ \hline
		\multirow{ 2}{*}{C-PR+Bi}  & 0.7904  &   0.7000   &  0.9233     
		& \multirow{ 2}{*}{0.9364} & \multirow{ 2}{*}{0.8971}  &  \multirow{ 2}{*}{0.9757}  &    0.7399  &  \textbf{0.5774} &  0.9186  \\
		& 0.7885* & 0.7014* & 0.9204*   &  &  & &   0.7512*  &  \textbf{0.5954}*     &  0.9218*    \\ \hline
	\end{tabular}
	\vspace{-0.25cm}
	\normalsize
\end{table*}

\begin{table*}[th]
\tiny
\centering
\caption{\small Comparisons of model interpretability of Top-$k$ features for C-PR and BB-PR on one Freebase and one ConceptNet test relation.}
\label{my-label}
\begin{tabular}{|p{8.25cm}|p{8.25cm}|}
\hline
\multicolumn{1}{|c|}{BB-PR}                                                      & \multicolumn{1}{c|}{C-PR}                                         \\ \hline
\multicolumn{2}{|c|}{Freebase}                                                                                                                          \\ \hline
\multicolumn{2}{|l|}{\textit{Test Relation}:~~\textbf{ /flim/film\_distributor/films\_distributed} ~~~~~~~~~~~~~ \textit{Avg. Precision}: [ BB-PR: \textbf{0.8050} ~vs.~ C-PR: \textbf{0.8906} ]~~~~~~ \textit{+ve F1 score}: [ BB-PR: \textbf{0.6667} ~vs.~ C-PR: \textbf{0.8232} ]} \\ \hline

\textcolor{blue}{award\_nominee$^{-1}\rightarrow$netflix\_genre/titles$^{-1}\rightarrow$netflix\_genres$^{-1}$}\textbf{;}~~ films\_distributed$\rightarrow$ film/distributors$\rightarrow$production\_company/films$\rightarrow$ film/release\_region$\rightarrow$film/release\_region$^{-1}$\textbf{;}~~ 
organization\_relationship/parent$\rightarrow$production\_companies$^{-1}$\textbf{;} \textcolor{blue}{location/mailing\_address/country} \textcolor{blue}{$\rightarrow$person/nationality$^{-1}\rightarrow$award\_winner$^{-1}\rightarrow$award\_nomination$^{-1}$}\textbf{;}~~ film/distributor$^{-1}\rightarrow$ film/release\_distribution\_medium$\rightarrow$film/release\_distribution\_medium$^{-1}\rightarrow$estimated\_budget/  currency$^{-1}\rightarrow$dated\_money\_value/currency$^{-1}$\textbf{;}~~ 
& 
films\_distributed$\rightarrow$nominated\_for$\rightarrow$film/starring $\rightarrow$flim/actor$\rightarrow$performance/film\textbf{;}~ organization\_relationship/child$\rightarrow$award\_winner$\rightarrow$films\_distributed\textbf{;}~
film\_distributor$^{-1}\rightarrow$ production\_companies$\rightarrow$films\_distributed\textbf{;}~~
film\_distributor$^{-1}\rightarrow$netflix\_genres$\rightarrow$production\_company/ films\textbf{;}~~ film\_distributor$^{-1}\rightarrow$film/distribution\_medium$\rightarrow$film\_distribution\_medium$^{-1}$\textbf{;}~ 
\\ \hline
\multicolumn{2}{|c|}{ConceptNet}                                                                                                                        \\ \hline
\multicolumn{2}{|l|}{\textit{Test Relation}:~~ \textbf{Antonym} ~~~~~~~~~~~~~~~~~\textit{Avg. Precision}: [ BB-PR: \textbf{0.6987} ~vs.~ C-PR: \textbf{0.7996} ]~~~~~~ \textit{+ve F1 score}: [ BB-PR: \textbf{0.4659} ~vs.~ C-PR: \textbf{0.6686} ]}                                                          \\ \hline
\textcolor{blue}{RelatedTo$^{-1}\rightarrow$RelatedTo$^{-1}$}\textbf{;}~ DerivedFrom$\rightarrow$Antonym$^{-1}$\textbf{;}~    Antonym$\rightarrow$Synonym\textbf{;}~   Antonym $\rightarrow$RelatedTo\textbf{;}~   \textcolor{blue}{EtymologicallyDerivedFrom$\rightarrow$EtymologicallyDerivedFrom$^{-1}$}\textbf{;}~ Antonym$^{-1}\rightarrow$ RelatedTo\textbf{;}~ \textcolor{blue}{IsA$\rightarrow$IsA$^{-1}$}\textbf{;}~ \textcolor{blue}{RelatedTo$\rightarrow$ RelatedTo$^{-1}\rightarrow$RelatedTo$^{-1}\rightarrow$RelatedTo}\textbf{;}~ Synonym $\rightarrow$ Antonym\textbf{;}~ \textcolor{blue}{EtymologicallyDerivedFrom$^{-1}\rightarrow$ RelatedTo}~ 
& Antonym$\rightarrow$Synonym$^{-1}$\textbf{;} Synonym$\rightarrow$Antonym$^{-1}$\textbf{;} \textcolor{blue}{IsA$\rightarrow$ RelatedTo$\rightarrow$RelatedTo$^{-1}$}\textbf{;} Antonym$^{-1}\rightarrow$Synonym$^{-1}$\textbf{;} \textcolor{blue}{RelatedTo$\rightarrow$IsA}\textbf{;} RelatedTo$\rightarrow$Antonym$^{-1}\rightarrow$RelatedTo\textbf{;} Antonym$\rightarrow$Synonym\textbf{;} \textcolor{blue}{EtymologicallyDerivedFrom$^{-1}$}\textbf{;} RelatedTo$\rightarrow$Antonym\textbf{;} Antonym$^{-1}\rightarrow$RelatedTo 
\\ \hline
\end{tabular}
\vspace{-0.3cm}
\normalsize
\end{table*}

\section{Experiments}
We evaluate C-PR empirically in terms of scalability, predictive performance and model interpretability. We performed experiments on Amazon EC2 linux server having Intel Xeon processor, 264GB RAM and 64 CPU cores in order to support the baselines that use exhaustive path search. 

\subsection{Data}
\label{sec:data}
We used three datasets for our experiments: two standard ones, viz (1) FB15k \footnote{https://everest.hds.utc.fr/doku.php?id=en:smemlj12}, (2) WordNet $^4$ and (3) one dataset created by us from ConceptNet 5 triples. FBK15k is a relatively dense subgraph of Freebase. The WordNet dataset is comparatively small and ConceptNet dataset is the largest one among the three. We choose these three datasets in order to evaluate C-PR on \textit{semantically different}\footnote{Freebase contains facts about persons, places and things; WordNet is a lexical KB of English; ConceptNet is a common sense KB.} KBs with \textit{diverse size} and \textit{density}. Using each dataset, we build a fairly large graph and use it as input to C-PR and the baselines. We also augment the graph with inverse relation triples ($t$, $r^{-1}$, $h$) for each ($h$, $r$, $t$) following the baseline methods. Table 2 shows the statistics of the graph and the labeled datasets. 

\subsection{Experimental Setup}
\label{sec:exp_setup}
\textbf{Labeled Dataset Generation.} For each of the three datasets, the triple number varies significantly across relations. Thus, to make the train and test data as \textit{balanced} as possible for all testing relations and to experiment with \textit{fairly large} labeled set of instances, we search for relations in our graph with $\geq$ 1000 instances\footnote{We observed that the number of triples for most relations are very small causing small train and test size after splitting. Choosing relations with triple number $\geq k$ as test relations enables creating balanced dataset of size \textit{at least} $k$ for all test relations.}. For Freebase, we found 86 such relations and randomly selected 25\footnote{Due to high resource usage, cost and training time for SFE-PR and SFE-PR+AR, it is not possible to evaluate all relations in KB.} \textit{semantically distinct} relations from various domains like film, location, music, sports, education etc. For each of the 25 relations, we randomly shuffle the list of distinct triples, choose 1000 triples and split them into 80\% training, 20\% test. As the number of relations in WordNet is relatively small, we select all 18 relations for test. For ConceptNet, we found 19 relations with $\geq$ 1000 instances and consider them for test. The train and test split for these two datasets is done in the same way as for Freebase.   

Given a relation $r$ and an observed triple ($h$, $r$, $t$) in training or testing, the entity-pair ($h$, $t$) is regarded as a positive instance for $r$. Following \cite{wang2016knowledge}, for each positive instance ($h$, $t$), we generate four negative ones, two by randomly corrupting the head $h$, and the other two by corrupting the tail $t$ by using only entities that have appeared in corresponding positions. Note that, the testing triples are not used for constructing the graph and also, none of the negative instances overlap with the positive ones.

\noindent \textbf{Baselines}.  We consider the following representative and state-of-the-art PR-based algorithms as baselines. We do not consider non-PR based methods here, as the path ranking (PR) algorithm outperforms embedding-based TransE model \cite{bordes2013translating} on FB15k \cite{wang2016knowledge}.

\textit{\textbf{Binarized Two-sided PR (B-PR)}} is the path-ranking algorithm described in Section \ref{PRA}. Following \cite{gardner2015efficient}, we also consider binarized path features for learning the PR-classifier like in our proposed C-PR.

\textit{\textbf{Binarized Bidirectional PR (BB-PR)}} is an extension of B-PR where we replace unidirectional two-sided RW (Random Walk) with bidirectional RW following the recent work \cite{lao2015learning}. BB-PR can be thought of as a version of C-PR that does not use context for path enumeration task. 

\textit{\textbf{B-PR+Bi}} is a simple extension to B-PR which augments the feature set with bigrams as in \cite{neelakantan2015compositional}.

\textit{\textbf{BB-PR+Bi}} is an extension to BB-PR with bigrams.

\textit{\textbf{SFE-PR\footnote{https://github.com/matt-gardner/pra}}} is a PR-based method that uses BFS (breadth-first search) with Subgraph Feature Extraction (SFE) for extracting path features \cite{gardner2015efficient}. SFE-PR uses only PR-style features for feature matrix construction. 

\indent \textit{\textbf{SFE-PR+AR$^8$}} is an extension to SFE-PR with ``\textit{Any-Relation}'' features \cite{gardner2015efficient}.

\textit{\textbf{DNA-PR}} is a modified version (to support path ranking) of the DNA algorithm (Sec. 2) where (1) we use the pre-trained word2vec models\footnote{The reference text corpus used by \cite{freitas2014distributional} for learning distributional semantics is not publicly available.} for learning distributional semantics like in C-PR and (2) employ unidirectional random walk (instead of unidirectional exhaustive search like in DNA) while exactly following the context-aware path finding strategy as proposed in DNA \cite{freitas2014distributional}. We use 0.05 as the similarity threshold as used in \cite{freitas2014distributional}.

\textit{\textbf{DNA-PR+Bi}} is an extension to DNA-PR with bigrams.

\noindent \textbf{Evaluation Metrics.} We use mean average precision (MAP), where p@k is computed following \cite{lao2011random}. We also use Avg. F1 score for +ve and -ve classes for evaluation. 

\subsection{Results and Analysis}
\label{sec:results}
We compare C-PR and baselines in three aspects as discussed below. The results for C-PR and random-walk based baselines are obtained using maximum path length $\eta$ = 7\footnote{Smaller $\eta$ ($\leq$ 4) often causes random walk to stop before reaching target. Larger $\eta$ ($\geq$10) reduces feature overlapping (due to specificity of longer paths) making feature matrix too sparse.} and $numWalkers$ = 20 \footnote{Using low value of $numWalkers$ (like 20) gives a small set of extracted features. This sets the stage for evaluating the main claim of C-PR: \textit{gaining higher predictive performance by learning from \textbf{fewer} number of discriminative path features.}}. In Tables 3 and 4, for SFE-PR+AR's results (marked *) due to extremely high RAM usage and training time ($> 6.5$ hrs), we had to terminate the execution of 2 ConceptNet and 6 Freebase test relations. 

\textbf{Evaluation-I: Scalability.} In this experiment, we evaluate C-PR and baselines in terms of Avg. number of features extracted for training, total RAM consumption in training all test relations and Avg. training time per test relation. From the results in Table 3, it is clear that SFE-PR and SFE-PR+AR end up extracting an exponentially large number of features (due to the use of exhaustive path search like BFS) which drastically increases their training time. Whereas, C-PR and other random-walk based baselines train on significantly fewer features which makes their training time significantly less. Moreover, the feature explosion problem in SFE-PR and SFE-PR+AR accounts for the extremely high RAM consumption which makes their execution infeasible with limited computational resources. For example, we could not train all test relations using SFE-PR on a local sever having 132GB of RAM in a single run. 
We noted that the highest RAM consumption for SFE-PR+AR during training a ConceptNet relation `\textit{causes}' with 2.25B features is 142GB and the corresponding training time is 6.25 hrs. 

\textbf{Evaluation-II: Predictive Performance.} Table 4 shows the comparative results of predictive performance of C-PR and baselines. To judge the overall improvements in predictive performance, we performed paired t-test considering +ve F1 scores (or avg. precision) on each relation as paired data.  For both Freebase and ConceptNet, C-PR outperforms all baselines with $p < 0.01$. For Freebase and WordNet, C-PR outperforms C-PR+Bi with $p < 0.05$. For ConceptNet, although C-PR+Bi outperforms C-PR with $p < 0.01$ based on Avg. +ve F1, the improvement of C-PR over C-PR+Bi considering MAP is statistically insignificant. In other words, augmenting C-PR with path bigram features doesn't ensure consistency in performance improvement. For WordNet, C-PR outperforms all baselines \textit{except} BB-PR with $p < 0.01$. The improvements of C-PR over BB-PR is statistically indistinguishable due to small size and sparsity of the WordNet dataset. From the analysis performed over three KBs, we can conclude that the \textit{denser} (like Freebase) and the larger (like ConceptNet) the KB becomes, the better the C-PR performs compared to the baselines, which is very important as real-life KBs are getting larger and larger. For smaller and sparser KBs like WordNet, C-PR performs as good as BB-PR. 

\textbf{Evaluation-III: Model Interpretablity.} Table 5 shows the results of model interpretability experiments performed following \cite{wang2016knowledge}. We selected top-$k$ features for one Freebase relation ($k=5$) and one ConceptNet relation ($k=10$) and compared C-PR with BB-PR (performed best among all baselines). We exclude WordNet from this analysis as the performance improvement of C-PR over BB-PR on Wordnet is statistically indistinguishable. Features marked with blue color in Table 5 represents less interpretable and black ones are more interpretable to human perception. For example, for relation `\textit{Antonym}', an interpretable feature for C-PR is ``\textit{Antonym}$\rightarrow$\textit{Synonym}$^{-1}$'' which can be explained as: if $a$ is antonym of $b$ and $b$ is synonym of $c$, then $a$ is antonym of $c$. Out of top-10 features for \textit{Antonym}, we see that 7 features are interpretable for C-PR and 5 features are interpretable for BB-PR. And for relation \textit{flims\_distributed}, all top-5 features are interpretable for C-PR whereas only 3 features are interpretable for BB-PR. This analysis suggests that, C-PR discovers more interpretable features than BB-PR as a result of its context-aware path search strategy, which also accounts for the gain in its predictive performance.  

\section{Conclusion}
\label{sec:length}
This paper proposed C-PR to extend path ranking algorithm by incorporating a context-aware path search strategy in random walk. Experimental results on three large-scale, real-world KBs show that the strategy enables discovery of much \textit{fewer}, more \textit{discriminative} and \textit{interpretable} path features compared to existing baselines, which in turn accounts for C-PR's significant gain in predictive performance with \textit{reasonable RAM consumption} and \textit{significantly less training time}. This makes C-PR a viable solution for performing inference on large-scale KBs with limited computational resources.

\section*{Acknowledgments} 
This work was supported in part by National Science Foundation (NSF) under grant no. IIS-1407927 and IIS-1650900. 

\bibliographystyle{named}
\bibliography{ijcai17}

\end{document}